\documentclass[natbib,smallcondensed]{svjour3}                     
\smartqed  
\usepackage{graphicx}
\usepackage{amssymb}
\usepackage[tbtags]{amsmath}
\usepackage{enumerate}
\usepackage{multirow}
\usepackage{verbatim}
\usepackage{epstopdf}
\usepackage{sidecap}
\usepackage{subfig}
\usepackage{graphicx, color}
\usepackage{caption}
\usepackage{amsmath}
\DeclareMathOperator*{\argmin}{argmin}
\DeclareMathOperator*{\argmax}{argmax}
\usepackage{algorithm2e}
\usepackage{float}
\usepackage{subfloat}
\usepackage{hyperref}
\usepackage{booktabs}

\journalname{Machine Learning}
\usepackage{subfig}
\usepackage{sidecap}

\begin{document}

\title{Deep Collective Matrix Factorization for Augmented Multi-View Learning}

\author {Ragunathan Mariappan \and Vaibhav Rajan }
\institute {School of Computing, National University of Singapore.
\email{vaibhav.rajan@nus.edu.sg}}
\date{}
\titlerunning{Deep Collective Matrix Factorization}

\maketitle
\begin{abstract}
Learning by integrating multiple heterogeneous data sources is a common requirement in many tasks. Collective Matrix Factorization (CMF) is a technique to learn shared latent representations from arbitrary collections of matrices. It can be used to simultaneously complete one or more matrices, for predicting the unknown entries. Classical CMF methods assume linearity in the interaction of latent factors which can be restrictive and fails to capture complex non-linear interactions. In this paper, we develop the first deep-learning based method, called dCMF, for unsupervised learning of multiple shared representations, that can model such non-linear interactions, from an arbitrary collection of matrices. We address optimization challenges that arise due to dependencies between shared representations through Multi-Task Bayesian Optimization and design an acquisition function adapted for collective learning of hyperparameters. Our experiments show that dCMF significantly outperforms previous CMF algorithms in integrating heterogeneous data for predictive modeling. Further, on two tasks -- recommendation and prediction of gene-disease association -- dCMF outperforms state-of-the-art matrix completion algorithms that can utilize auxilliary sources of information.
\end{abstract}

\section{Introduction}


Pairwise relational data, found in many domains, can be represented as matrices. 
Matrix completion, that predicts unknown entries in a matrix, is widely used in many applications, e.g. in
recommender systems \citep{koren2009matrix}, 
computer vision \citep{hu2012fast} and
bioinformatics \citep{natarajan2014inductive}, 
to name a few.
Often, the matrices are high-dimensional, sparse,
and with inherent redundancies. Sufficient information may be present in latent substructures, that can be approximated through
low-rank factorizations, and used in predictive models.

\begin{figure*}[!t] 
\centering
        \subfloat[]{
        \centering
                \fontsize{8pt}{8pt}\selectfont
                \includegraphics[width=0.25\textwidth]{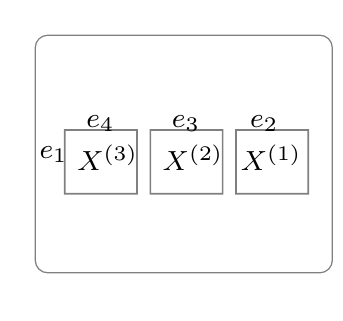}
                \label{eg:a}       
        }
        \subfloat[]{
        \centering
                \fontsize{8pt}{8pt}\selectfont
                \includegraphics[width=0.25\textwidth]{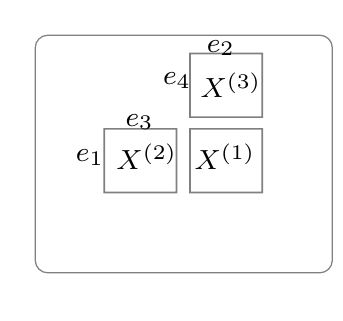}
                \label{eg:b}           
        }
        \subfloat[]{
        \centering
                \fontsize{8pt}{8pt}\selectfont
                \includegraphics[width=0.25\textwidth]{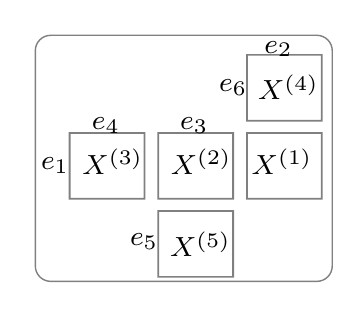}
                \label{eg:c}
        }
        \caption{Examples of 
        (a) \textbf{multi-view} setting, 
        (b) \textbf{recommendation} setting: 4 entities $e_1, e_2, e_3, e_4$ and 3 relations between the entities, matrices $X^{(1)},X^{(2)},X^{(3)}$; 
        (c) \textbf{augmented multi-view} setting: 6 entities $e_1, e_2, e_3, e_4, e_5, e_6$ and 5 relations between the entities, matrices $X^{(1)},X^{(2)},X^{(3)},X^{(4)},X^{(5)}$.}
        \label{eg}
\end{figure*}

When information from multiple heterogeneous sources is available, predictive models benefit from 
latent representations that model correlated shared structure.
In multi-view learning, {\it views} refer to measurements for the same subjects, that differ in source, datatype or modality.
Each matrix, representing a view, has a relationship between two {\it entity types}, along each matrix dimension, and entity types may be involved in multiple views.
For example, in fig. \ref{eg}(a), entity $e_1$ could be patients and clinical data from three different sources 
(notes $X^{(1)}$, images $X^{(2)}$, and diagnoses $X^{(3)}$) may be used to obtain patient representations for modeling risk of diseases.
When auxiliary information about multiple entity types are present, they could be effectively utilized to obtain latent representations. For example, in hybrid recommender systems, where side information matrices about users and movies are used in addition to the historical user-rating matrix to obtain user and movie representations (in fig. \ref{eg}(b), $X^{(1)}$ is the user-rating matrix, $X^{(2)}$ has user-features and $X^{(3)}$ has movie-features). These latent representations are then used to recommend movies to users.

{\it Collective Matrix Factorization (CMF)} is a general technique to learn shared representations from arbitrary collections of heterogeneous data sources \citep{singh2008relational}.
CMF collectively factorizes the input set of matrices to learn a low-rank latent representation for each entity type from {\it all} the views in which the entity type is present.
It can be used to
simultaneously complete one or more matrices in the collection of matrices.
Since CMF models arbitrary collections of matrices, this setting is also referred to as {\it augmented multi-view learning} \citep{klami2013group}.
Fig. \ref{eg}(c) shows an example. Note that the augmented multi-view setting can generalize to any collection of matrices and subsumes the multi-view and recommendation settings. 

Classical matrix factorization based approaches assume linearity in the interaction of latent factors which can be restrictive and fails to capture complex non-linear interactions.
Modeling such non-linearities through neural models 
have significantly improved multi-view learning approaches 
with two views \citep{andrew2013deep,wang2015deep} and multiple (but not augmented) views \citep{ngiam2011multimodal,wang2017multi}.
A common approach is the use of deep  autoencoders to obtain shared representations that form latent factors.
However these methods cannot generalize to arbitrary collections of matrices. 
To use shared representations from autoencoders within CMF, learning would involve optimizing entity-specific autoencoder reconstruction losses as well as view-specific matrix reconstruction losses.
The latter induces dependencies between the autoencoder networks that may result in simultaneous under-fitting in some networks and over-fitting in other networks 
(described in section \ref{optimization}).
This makes collective learning of all latent representations challenging and, to scale to arbitrary collections of matrices,  
necessitates automatic hyperparameter selection.



In this paper,
we develop dCMF, a deep learning architecture for collective factorization of arbitrary collection of matrices, that is, to our knowledge, the first deep augmented multi-view learning method. dCMF overcomes the limitation of previous CMF models that cannot capture complex non-linear interactions of latent factors.
We address optimization challenges that arise due to dependencies between autoencoder representations within dCMF, through multi-task Bayesian optimization and an acquisition function that is adapted for collective learning of hyperparameters.
Our experiments show that dCMF is better than previous CMF algorithms at integrating heterogeneous data for predictive modeling and significantly outperforms them on synthetic and real data.
We demonstrate two applications of dCMF in matrix completion tasks: movie recommendations and prediction of gene-disease associations. In both tasks, dCMF significantly outperforms state-of-the-art algorithms on benchmark datasets.


\vspace{-1.25em}
\section{Related Work}


\noindent
{\bf Multi-View Learning.}
Canonical Correlation Analysis (CCA) \citep{hotelling1936relations} that learns maximally correlated features from two views has been the basis for many multi-view learning methods.
Several variants have been studied that illustrate the benefits of multi-view learning over models that learn from concatenation of features in the views \citep{hardoon2004canonical}.
{\it Augmented Multi-View Learning} generalizes the setting to arbitrary collections of matrices where latent factors are learnt through collective matrix factorization (CMF) \citep{singh2008relational}, thus enabling learning from auxiliary data sources.
To model view-specific noise and allow a subset of matrices with shared structure independent of others, a group-wise sparse formulation was designed in gCMF \citep{klami2013group}.\\

\noindent
{\bf Deep Models for Multi-View Learning.}
Deep learning based extensions of some of these models have been developed.
Multi-modal autoencoders were designed to learn shared representations from multiple views, called SplitAE \citep{ngiam2011multimodal}.
DCCA is a deep extension of CCA that maximizes correlation between non-linearly extracted features from each view \citep{andrew2013deep}.
DCCAE combines DCCA and SplitAE to get shared representations by maximizing a CCA-based objective  \citep{wang2015deep}.
Thus DCCAE is designed for a multi-view setup with 2 views and 3 entities (e.g., only views $X^{(1)}, X^{(2)}$ in fig. \ref{eg}(a)) and 
obtains shared representations by maximizing the canonical correlation between the unshared entities ($e_2, e_3$), regularized by the autoencoder reconstruction error. 
All these deep models benefit from the non-linearities captured through the deep representations but are restricted to two input views.

There are supervised learning methods that model multi-view data. For example, 
in CDMF \citep{wang2017multi}, a deep learning based solution is developed for the multi-view case (not augmented multi-view) where each view differs in its modality.
They factorize each view into a modality-invariant factor and a modality-specific factor, where the latter is learnt using a neural network. 
Being supervised, entity representation learning in these methods is guided by application-specific labels.
None of these deep learning approaches can be used to model an arbitrary collection of matrices for unsupervised augmented multi-view learning.

\noindent
{\bf Heterogeneous Information Networks.}
Another approach to representation of (augmented) multi-view data is through Heterogeneous Information Networks (HIN).
A HIN contains multiple types of nodes (entities) and multiple types of edges (relations between entities).
HIN embeddings obtain vectorial representations of nodes that preserve global structural properties of the network.
Such embeddings can then be used for link prediction, node classification or clustering.
They have also been used for recommendation (e.g., by \cite{han2018aspect,shi2019heterogeneous}).
There are several approaches to learn HIN embeddings, including some that use deep neural networks \citep{chang2015heterogeneous}.
HIN models multiple relations between the same entities elegantly through multiple edge types while CMF-based methods would require specifying a matrix-specific link functions to model such relations.
However, integrating side information such as node and edge attributes in HIN embeddings is challenging \citep{cui2018survey}.
In contrast, matrix-based approaches naturally model edge attributes (as matrix entries) of any type -- binary, ordinal or real-valued. More details can be found in recent surveys \citep{shi2017heterogeneous,cui2018survey}.

\section{Background}
{\bf Matrix Factorization.} For a matrix $X \in \mathbb{R}^{m \times n}$, a low-rank factorization aims to obtain latent factors $U^{(1)} \in \mathbb{R}^{m \times K}, U^{(2)} \in \mathbb{R}^{n \times K}$, such that $X \approx U^{(1)} \cdot U^{{(2)}^T}$, where the $K < \min(m,n)$ (see fig. \ref{fig:dcmf_overview}\subref{fig:dcmf_overview:a}).
The factors are learnt by solving the optimization problem: $\argmin_{U,V} L(X, U^{(1)} \cdot U^{{(2)}^T)}$, where $L$ denotes a loss function (e.g., $|| X - U^{(1)} \cdot U^{{(2)}^T}||^2_F$, where $||.||^2_F$ is the Frobenius norm). 
Collaborative Filtering, for recommendations, uses such an approach where $X$ is the rating matrix. 
A common approach to solving this is through a convex relaxation that minimizes the nuclear norm (the sum of singular values) of 
$U^{(1)} \cdot U^{(2)^T}$, which is equivalent to solving: $\min_{X = U^{(1)} \cdot U^{(2)^T}} ||U^{(1)}||^2_F + ||U^{(2)}||^2_F$ \citep{srebro2005rank}.\\

\noindent
{\bf CMF} 
aims to jointly obtain low-rank factorizations of 
$M$ matrices (indexed by $m$),  $X^{(m)}  = [x_{ij}^{(m)}]$, that describe relationships between $E$ entities ($e_1,\dots e_E$), each with dimension $d_e$.
The entities corresponding to the rows and columns of the $m^{th}$ matrix are denoted by $r_m$ and $c_m$ respectively.
Fig. \ref{eg} shows three examples.
Each matrix is approximated by product of low rank-$K$ factors that form the representations of the associated row and column entities:
$X^{(m)} \approx U^{(r_m)} U^{{(c_m)}^T} $
where $U^{(e)} = [u_{ik}^{(e)}] \in \mathbb{R}^{d_e \times K}$ is the low-rank matrix for entity type $e$. 
Any two matrices sharing the same entity type use the same low-rank representations as part of the approximation, which enables sharing information.
For example, in fig. \ref{eg}\subref{eg:b}, the same latent factor $U^{(e_1)}$ is used to reconstruct the three matrices $X^{(1)} \approx U^{(e_1)}  U^{{(e_2)}^T}, X^{(2)} \approx U^{(e_1)}  U^{{(e_3)}^T}, X^{(3)} \approx U^{(e_1)} U^{{(e_4)}^T}$.
The latent factors are learnt by solving the optimization problem: 
\begin{equation}\label{eq:CMFLoss}
\argmin_{\{U^{(e)} \in \mathbb{R}^{d_e \times K}\}_{e \in E}}\,\,\, \sum_{m=1}^M L(X^{(m)}, U^{(r_m)} U^{{(c_m)}^T}) + \sum_{e=1}^E \mathcal{R}(U^{(e)}) 
\end{equation}
where $M$ is the total number of input matrices, $E$ is the total number of entities and $\mathcal{R}$ is a regularizer.
For $\mathcal{R}(U^{(e)}) = \lambda ||U^{(e)}||^2_F$, \cite{bouchard2013convex} show that this formulation generalizes the nuclear norm for a single matrix to a {\it collective nuclear norm} defined on an arbitrary set of matrices (with the reasonable assumption that a pair of entity types do not share more than one view). Although this is a non-convex problem, in practice, solutions obtained through Stochastic Gradient Descent yield good performance \citep{bouchard2013convex}.




\section{Problem Statement}
Given $M$ matrices (indexed by $m$),  $X^{(m)}  = [x_{ij}^{(m)}]$, that describe relationships between $E$ entities ($e_1,\dots e_E$), each with dimension $d_e$,
we aim to jointly obtain latent representations of each entity $U^{(e_i)}$ and low-rank factorizations of each matrix $X^{(m)} \approx U^{(r_m)} \cdot U^{{(c_m)}^T}$,
such that $U^{(e_i)} = f^{e_i}_{\theta}([X]_{e_i})$
where $f^{(.)}$ is an entity-specific non-linear transformation parameterized by $\theta$ and $[X]_{e_i}$ denotes all matrices in the collection that contains a relationship of entity $e_i$.
The entities corresponding to the rows and columns of the $m^{th}$ matrix are denoted by $r_m$ and $c_m$ respectively.

\begin{figure*}[h]
       \centering 
        \subfloat[]{
                \centering
                \fontsize{7pt}{7pt}\selectfont
                \includegraphics[width=0.45\textwidth]{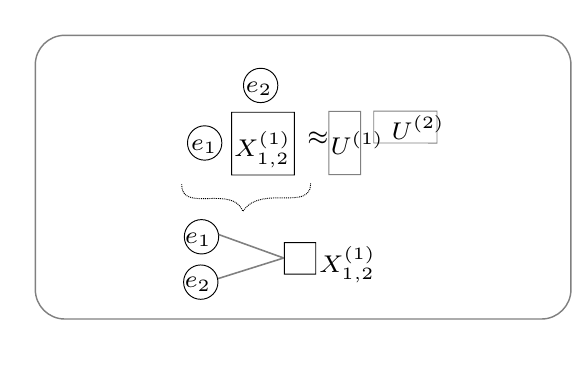}
                \label{fig:dcmf_overview:a}
        }
        \subfloat[]{
                \centering
                \fontsize{7pt}{7pt}\selectfont
                \includegraphics[width=0.495\textwidth]{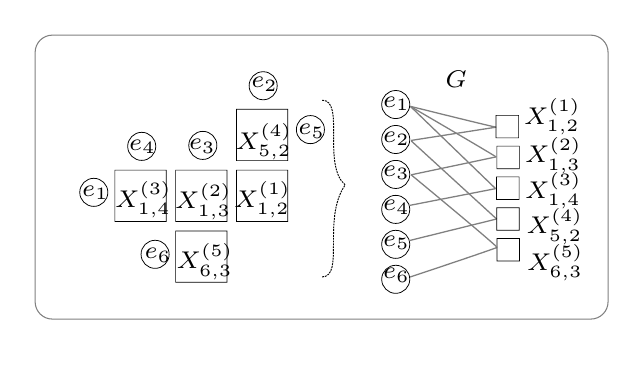}
                \label{fig:dcmf_overview:b}
        } \\
        \subfloat[]{
                \centering
                \fontsize{7pt}{7pt}\selectfont
                \includegraphics[width=0.975\textwidth]{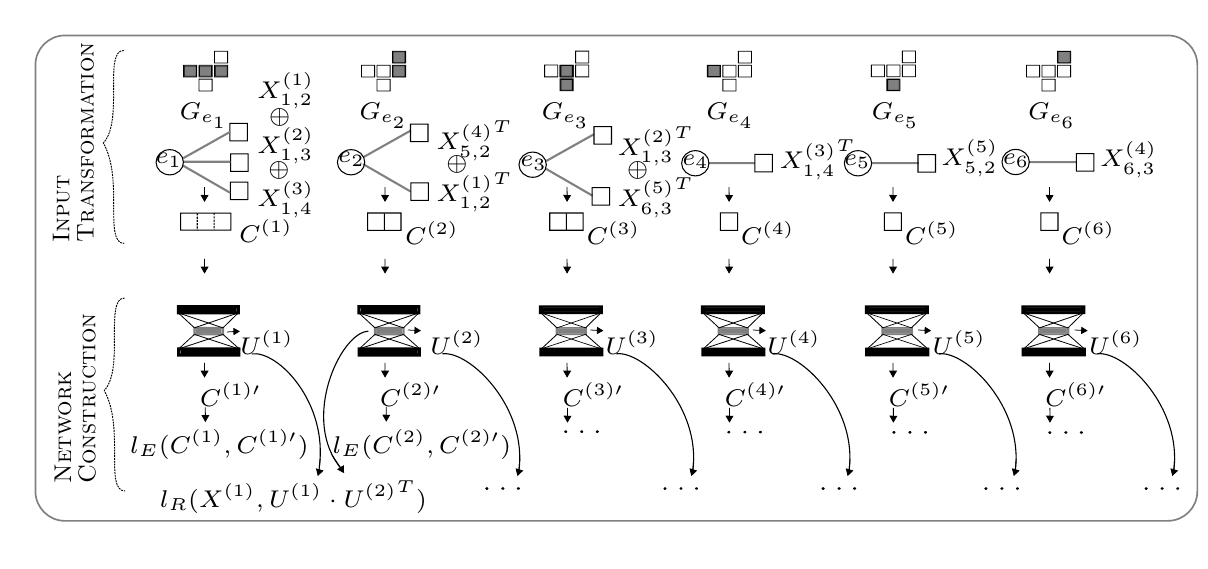}
                \label{fig:dcmf_overview:c}
        }
        \caption{(a) Entity-matrix relationship graph for a single view (b) A collection of views and its entity-matrix relationship graph [square nodes: matrices, circular nodes: entities] (c) dCMF model construction for the example in fig. (b).}\label{fig:dcmf_overview}
\end{figure*}

We assume that the relationship between these matrices and the constituent $E$ entities is
provided as a
bipartite entity-matrix relationship graph $G(V_E, V_M, D)$, where vertices $V_E, V_M$ represent entities and matrices respectively. 
Edges $(e_i,X^{(m)})$, $(e_j,X^{(m)})  \in D$ are present if there exists an input matrix $X^{(m)} \in V_M$ capturing the relationship between the entities $e_i, e_j \in V_E$ (see fig. \ref{fig:dcmf_overview}\subref{fig:dcmf_overview:a},\subref{fig:dcmf_overview:b}).

\section{Deep Collective Matrix Factorization (dCMF)}
Similar to the formulation in (\ref{eq:CMFLoss}), we aim to learn 
entity-specific latent representations
by solving the following optimization problem:
\begin{equation}\label{eq:dCMFLoss}
\argmin_{\theta, \{U^{(e)} \in \mathbb{R}^{d_e \times K}\}_{e \in E}}\,\,\, \sum_{m=1}^M L(X^{(m)}, f^{r_m}_{\theta}([X]_{r_m})
 \cdot (f^{c_m}_{\theta}([X]_{c_m}))^T) + \sum_{e=1}^E \mathcal{R}(U^{(e)}) 
\end{equation}
We now describe how we model $f$ to induce non-linearity and $\mathcal{R}$ for regularization, to develop the dCMF model.

There are several ways to model non-linearity ($f$).
Common choices in unsupervised learning include kernels, 
Restricted Boltzmann machine (RBM) \citep{hinton2006reducing}
and Autoencoders \citep{ngiam2011multimodal}. 
Kernel machines are not recommended for representation learning due to its over-reliance on the smoothness assumption,
i.e., the value of the learned function at a data point depends mostly on the training examples that are closest to it
\citep{bengio2013representation}. 
Instead \cite{bengio2013representation} advocate nonparametric models such as neural networks whose model complexity can be controlled through hyperparameters. 
RBM is a stochastic generative model with intractable maximum likelihood function and complex training procedures. 
In contrast, autoencoders can be trained efficiently and 
have been used effectively for multi-view representation learning \citep{wang2015deep}.
The use of autoencoders to model $f$ also 
allows us to use the autoencoder reconstruction loss as the regularizer $\mathcal{R}$. 
Such autoencoder-based regularization has been used in DCCAE \citep{wang2015deep} for multi-view learning from two views.
There are several autoencoder architectures \citep{goodfellow2016deep} that can be used; we leave that investigation for future work. Here we choose the simplest architecture with multiple fully-connected hidden layers.

Since entities can be shared across matrices, we have to obtain autoencoder-based shared representations from multiple matrices simultaneously to enable sharing of information across matrices. 
Further, this must be done in a way that can be generalized to an arbitrary collection of input matrices.
To accomplish this we design a neural architecture as described in the following section.
Generalizing to augmented multi-view learning, compared to relatively simpler settings like  multi-view learning (fig. \ref{eg:a}), leads to non-trivial optimization challenges that we discuss and address in later sections.

\subsection{Model construction and training}

There are two steps in dCMF model construction:
\begin{enumerate}
\item 
Input Transformation:
For each entity $e_i$, we create a new matrix $ C^{(i)}$, that we call {\it concatenated matrix}, by concatenating all the matrices containing entity $e_i$, i.e., all the matrices that are neighbors of $e_i$ in $G(V_E,V_M,D)$. 
Note that we transform $M$ input matrices to $E$ concatenated matrices, and a single input matrix ($X^{(m)}$) may be used in multiple concatenated matrices ($C^{(i)}$).
The concatenation ensures that for each entity, we use the information from all available input matrices to learn its representation.
\item
Network Construction:
We then use $E$ (dependent) autoencoders to obtain the latent factors $U^{(i)}$ from the concatenated matrices $C^{(i)}$.
For each entity $e_i$ our network has an autoencoder whose input is $C^{(i)}$, and the decoding is represented by $C^{(i)^\prime}$.
The bottleneck or encoding of each autoencoder, after training, forms the latent factor $U^{(i)}$.
The latent factors are learnt by training all the autoencoders together by solving:
\begin{equation}\label{eq:dCMF}
    \argmin_{\{U^{(e)} \in \mathbb{R}^{d_e \times K}\}_E}\,\,\,  \sum_{m=1}^{M} l_R(X^{(m)},X^{(m)\prime})) + \sum_{e=1}^E l_{E}(C^{(e)}, C^{(e)\prime}) 
\end{equation}
where $l_E$ is the reconstruction loss between the autoencoder's input $C^{(i)}$ and the decoding $C^{(i)\prime}$; $l_R$ is the matrix reconstruction loss, where the reconstructed matrix $X^{(m)\prime}=U^{(r_m)} \cdot U^{{(c_m)}^T}$ of the view $X^{(m)}$ is obtained by multiplying the associated row and column entity representations $U^{(r_m)}$ and $U^{(c_m)}$.
We call the summations in equation (\ref{eq:dCMF}) the matrix reconstruction loss ($\mathcal{L}_R$) and autoencoder reconstruction loss ($\mathcal{L}_E$) respectively.
\end{enumerate}

Thus, while CMF factorizes each matrix as $X^{(m)} \approx U^{(r_m)} \cdot U^{{(c_m)}^T}$,  dCMF performs non-linear factorization using autoencoders as  $X^{(m)} \approx g_{\theta}^{(r_m)}(C^{(r_m)}) \cdot g_{\theta}^{(c_m)}(C^{{(c_m)}^T})$, where $g_{\theta}$ is the encoder corresponding to the entity, with parameter set $\theta$, obtained by collectively minimizing the sum of all the matrix reconstruction and autoencoder reconstruction losses as described above.\\

\noindent
{\bf Illustration.} 
In fig. \ref{fig:dcmf_overview}\subref{fig:dcmf_overview:a} we show a single matrix $X^{(1)}_{1,2}$ and its two entities $e_1$ and $e_2$. The corresponding entity-matrix graph below has 2 circular nodes for two entities and 1 square node for the matrix.
In fig. \ref{fig:dcmf_overview}\subref{fig:dcmf_overview:b}, we show the graph for the collection of 5 matrices and 6 entities ($e_1$ to $e_6$) (from fig. \ref{eg}(c)). 
Consider, for instance, the entity $e_1$. There exists 3 matrices with relationships of entity $e_1$ with three other entities $e_2$, $e_3$ \& $e_4$. Hence there are 3 edges from the node representing $e_1 \in V_E$ to the nodes $X^{(1)},X^{(2)},X^{(3)} \in V_M$. 

We illustrate dCMF model construction in fig. \ref{fig:dcmf_overview}(c) for the example from fig. \ref{fig:dcmf_overview}(b). 
We construct $E=6$ autoencoders, one per entity. 
The autoencoder construction
for entity $e_1$ is illustrated in the first column of fig. \ref{fig:dcmf_overview}(c). We show the subgraph $G_{e_1}$ consisting of 3 edges corresponding to the 3 views $X^{(1)}_{1,2}$, $X^{(2)}_{1,3}$, $X^{(3)}_{1,4}$. 
Hence $C^{(1)}$ = $X^{(1)}_{1,2} \oplus X^{(2)}_{1,3} \oplus X^{(3)}_{1,4}$, where $\oplus$ denotes row or columnwise concatenation of the matrices. 
To pictorially illustrate this we show a miniature of the setup in fig. \ref{fig:dcmf_overview}(b) on top of each column in fig. \ref{fig:dcmf_overview}(c), greying out the boxes corresponding to the matrices involved in $C^{(i)}$ construction. We also show $C^{(i)}$ as a block containing concatenated boxes (equal to the number of matrices $C^{(i)}$ is composed of) with a label $C^{(i)}$ below each of the subgraphs which is also the input to the corresponding autoencoder. 
Similarly we construct the autoencoder for $e_2$ and the input $C^{(2)}$ by concatenating matrices corresponding to the edges of $G_{e_2}$ as illustrated in the second column of fig. \ref{fig:dcmf_overview}(c). Thus we have 6 columns in fig. \ref{fig:dcmf_overview}(c) for each of the 6 entities in setup of fig. \ref{fig:dcmf_overview}(b). 
To avoid clutter in fig. \ref{fig:dcmf_overview}(c), we show only two examples of the autoencoder reconstruction loss $l_E$ for entities $e_1$ and $e_2$ and one example matrix reconstruction loss $l_R$ for the matrix $X^{(1)}$.
In total there are $E=6$ autoencoder reconstruction loss terms and $M=5$ matrix reconstruction loss terms.
Note that this construction can be generalized to any number of entities and matrices.

Notice that the input dimension, which depends on $C^{(e)}$, is different for each autoencoder and the bottleneck layer dimension (the chosen low rank $K$) is common across all autoencoders. 
So, the number of layers for each autoencoder is a hyperparameter that is chosen adaptively for each autoencoder as 
follows: We start with the autoencoder's input dimension obtained from $C^{(i)}$ and multiply it with a fraction $f_k$ (a hyperparameter) to get the size of the first encoding layer. We then multiply the first encoding layer's size again with $f_k$ to get the second encoding layer's size. We repeat this and continue to add layers until we cross $K$ which is the common encoding/bottleneck size for all the autoencoders. We add a decoding layer corresponding to each of the encoding layer. This approach helps to adaptively decide the number of layers and their size for each autoencoder based on their input size i.e. more layers are added for inputs of higher dimension and vice-versa.

\subsection{Optimization}\label{optimization}

dCMF learns the representation of all the input entities collectively by training all $E$ autoencoders simultaneously. The objective function  $\mathcal{L}_R + \mathcal{L}_E$ is non-convex. Note that 
although the autoencoder reconstruction losses in $\mathcal{L}_E$ are independent to each of the autoencoders, the matrix reconstruction losses in $\mathcal{L}_R$ are dependent on multiple autoencoders, through the latent factors used in matrix reconstruction.
Below we describe the problems that arise specifically due to
the augmented multi-view setup and how we address them.\\

\begin{figure*}[t] 
\centering
        \subfloat[]{
        \centering
                \fontsize{5pt}{5pt}\selectfont
                \includegraphics[width=0.475\textwidth]{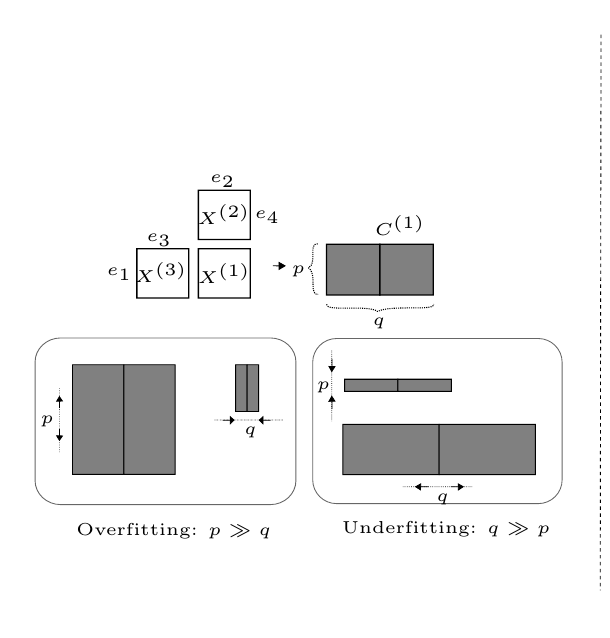}
            \label{fig:opt_probs:over-under-pb-illust}       
        }
        \subfloat[]{
        \centering
                \fontsize{5pt}{5pt}\selectfont
                \includegraphics[width=0.475\textwidth]{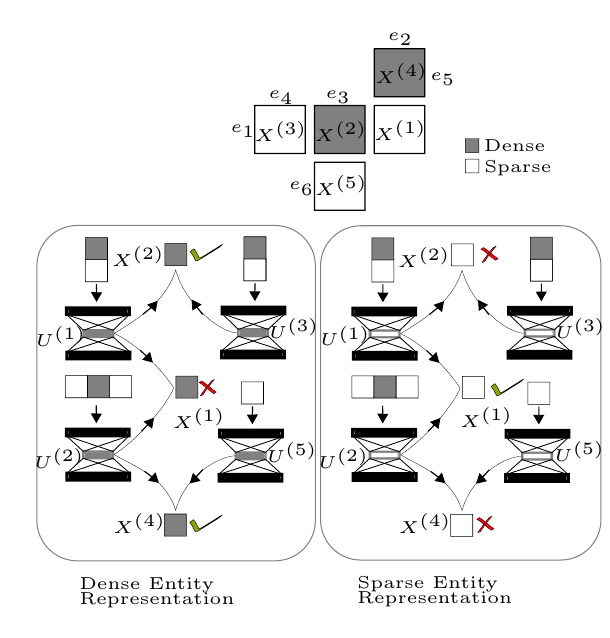}
                \label{fig:opt_probs:sparse-pb-illust}       
        }
        \caption{(a)  Entity size, shape and interactions in dCMF (b) Effect of mixed sparsity levels in dCMF.}
        \label{fig:opt_probs}
\end{figure*}

\noindent
{\bf Entity size, shape and interactions. }
The \textit{entity-size} $d_{e_i} = |e_i|$ is the number of instances of an entity $e_i$ i.e. count of rows/columns depending on whether it is the row/column entity of the matrix. Recall that the \textit{concatenated matrix} $C^{(i)}$ is constructed by concatenation of all the matrices $X^{(m)}$ associated with entity an $e_i$. The \textit{entity-shape} of an entity $e_i$ is $(p,q)$, where $p$ and $q$ are the row and column dimensions of the corresponding matrix $C^{(i)}$ respectively. Thus by virtue of dCMF's model construction $p$ and $q$ becomes the size and feature dimension of the input to the autoencoder for learning the entity representation. Let \textit{entity-interactions} $N_{e_i}$ be the number of matrices sharing entity $e_i$.
E.g., consider the entity $e_1$ in fig. \ref{fig:opt_probs}\subref{fig:opt_probs:over-under-pb-illust}. The matrix $C^{(1)}$ is constructed by row-wise concatenation of $X^{(1)}_{1,2}$, $X^{(3)}_{1,3}$. Here $q = |e_2| + |e_3|$, $p = |e_1|$ and $N_{e_1}=2$. 

If $q \gg p$ for an entity $e$ then there may be under-fitting in the learning of $U^{(e)}$, since the dimensionality of the autoencoder input $C^{(i)}$ is high and the number of samples are few. The situation worsens if $N_e$ is large i.e. the entity is related to many other entities, leading to increasing dimensions in $C^{(i)}$. Pre-training or increasing the number of hidden layers of the corresponding autoencoder may help in such conditions.
On the other hand, if $p \gg q$ then it may lead to over-fitting, since the dimensionality of $C^{(i)}$ is low and the number of samples are high. This may happen when the \textit{entity-size} of each of the associated $N_e$ entities are all small. This can be addressed by adding suitable regularizers or through early stopping during the training of the corresponding autoencoder.
A schematic is shown in fig. \ref{fig:opt_probs}\subref{fig:opt_probs:over-under-pb-illust}.
Thus, $|e|, p, q$ and $N_e$ can all influence dCMF performance
and require careful hyperparameter selection, separately for each autoencoder, 
noting that in dCMF, the autoencoder losses are not independent since  
$\mathcal{L}_R$ depends on multiple autoencoders collectively.
Since manually tuning these hyperparameters is infeasible for arbitrary collection of matrices, we address this problem through Bayesian Optimization.

\noindent
{\bf Mixed sparsity levels.} 
An augmented multi-view setup may contain both sparse and dense matrices. E.g., in recommendation, the rating matrix is sparse but side information matrices may be dense.
If an autoencoder's input is a concatenation of both sparse and dense matrices, then the learnt representation will be dense with potentially many small values in order to bring down the autoencoder reconstruction loss $l_E$. This results in a higher matrix reconstruction error $\mathcal{L}_R$ for the sparse matrices as the corresponding reconstructed matrices using dense row and column entity representations are not sparse. 
To handle this one can attempt to learn sparse representations using sparse autoencoders. But any dense matrix associated with the entity with sparse representation will suffer from higher $\mathcal{L}_R$. 
We illustrate this in fig. \ref{fig:opt_probs}\subref{fig:opt_probs:sparse-pb-illust} for the example in fig. \ref{fig:dcmf_overview}(b) with sparse $X^{(1)}$, $X^{(3)}$, $X^{(5)}$ and dense $X^{(2)}$, $X^{(4)}$. We can see that if the learnt representations for $U^{(.)}$ are dense then it favours reconstruction of $X^{(2)}$ and $X^{(4)}$ but not $X^{(1)}$. On the other-hand if the  representations learnt are sparse then it favours the reconstruction of $X^{(1)}$ but not $X^{(2)}$ and $X^{(4)}$.


Other autoencoder-based unsupervised deep learning algorithms do not face these challenges. For instance,
the Improved Deep Embedded Clustering (IDEC) algorithm \citep{guo2017improved} aims to learn entity representations that favour clustering, using autoencoders, by optimizing a clustering oriented loss regularized by the autoencoder reconstruction loss.
Although the objective function is similar, since they have a single autoencoder, problems due to multiple dependencies (through matrix reconstruction loss in dCMF) do not arise.

\subsubsection*{Hyperparameter tuning}

Manual tuning is infeasible for complex models with large number of hyperparameters and complex models; and random search or grid search based approaches \citep{bergstra2012random} are either too time-consuming or not effective.
In the case of dCMF, there are many hyperparameters related to (a) Model construction and (b) Optimization.
Hence dCMF training devoid of hyperparameter tuning usually results in poor performance due to the dependencies that may result in simultaneous over/under-fitting in different autoencoders as described earlier.
See Appendix \ref{ap:hyp} for a list of hyperparameters.

Hyperparameter tuning can be formulated as an optimization problem:
\begin{eqnarray}\label{hyp}
\textbf{p}^{*} = \underset{\textbf{p} \in \mathcal{P}}{\argmin}\,\,   \textbf{L}(\textbf{p})
\end{eqnarray}
where $\textbf{p}^{*}$, from the $B$-dimensional hyperparameter space $\mathcal{P}$ denotes the optimal values for all  $B$ hyperparameters and the objective function is the collective dCMF loss $\textbf{L} = \mathcal{L}_E+\mathcal{L}_R$. Note that model training optimizes $\textbf{L}$ with respect to the model parameters, and not the hyperparameters. 
The functional form of $\textbf{L}(\textbf{p})$ is not known but its value, for any given input $\textbf{p}$, can be computed. 
Thus, this is a black-box function optimization problem that has been successfully addressed through Bayesian Optimization \citep{snoek2012practical}.\\

\noindent
{\bf Multi-Task Bayesian Optimization (MTBO).}
We briefly explain Multi-Task Bayesian Optimization (MTBO) (for more details see \citep{bergstra2011algorithms,snoek2012practical,swersky2013multi}) before describing how dCMF adopts MTBO. 

Bayesian Optimization (BO) is a sequential model-based approach for solving the black-box function optimization problem. 
The key idea is to learn a {\it surrogate model} $\mathcal{M}$ that captures our beliefs about the unknown objective function ($\textbf{L}(\textbf{p})$).
This model is learnt from {\it data}, $\mathcal{D}_n$ = ${(\textbf{p}_1,\textbf{L}_1),\ldots,(\textbf{p}_n,\textbf{L}_n)}$, that consists of sequential evaluations of $\textbf{L}(\textbf{p})$ for different values of $\textbf{p}$.
Generating this data sequence requires making the decision of which $\textbf{p}$ to evaluate next, at each step.
This decision is made through an {\it acquisition function} $\alpha$.
These functions are designed to have optima at points with high uncertainty in the surrogate model (thus facilitating exploration) and/or at points with high predictive values in the surrogate model (thus facilitating exploitation).
Acquisition functions have known functional forms and are usually easier to optimize than the original objective function.
The surrogate model is updated sequentially with each observed data point.
Over multiple steps, the landscape of the black-box function ($\textbf{L}(\textbf{p})$) is learnt by the surrogate model and can be exploited by the acquisition function to yield values of $\textbf{p}$ that are, on average, closer to the optimal $\textbf{p}^*$.

Many different choices of surrogate models and acquisition functions have been explored. 
Gaussian processes (GP) can be used to model priors over functions and are closed under sampling which makes them an elegant choice for a surrogate model in BO -- after each data point is generated (using the acquisition function), the updated model is also a GP with updated mean and covariance functions. They have been successfully used in BO for hyperparameter optimization \citep{bergstra2011algorithms,snoek2012practical}.

 \begin{algorithm}[t]
    \SetKwInOut{Input}{inputs}
    \SetKwInOut{Output}{outputs}
    \SetKwProg{BO}{BO}{}{}
    \BO{($\mathcal{N},\textbf{L})$}{
        \Input{\text{dCMF Network}, $\mathcal{N}$ and Loss function, \textbf{L}} 
        \Output{Best hyperparameter set $\textbf{p}^\#$ and trained network ($\mathcal{N}$) parameters $\Theta^\#$ } 
        Generate Data $\mathcal{D}_m$ = ${(\textbf{p}_i,\textbf{L}_i,V_i,\Theta_i),\,\, i = 1,\ldots,m}$ from $m$ random samples of $\textbf{p}$ \\
         \quad with corresponding validation set performance: $V_1,\dots V_m$\\ \quad and network parameters: $\Theta_1,\dots \Theta_m$.\\
        Train MTGP surrogate model on $\mathcal{D}_m$: $\mathcal{M}(\boldsymbol{\mu},\boldsymbol{\sigma};\mathcal{D}_m(\textbf{p},\textbf{L}))$\\
         \For{n = m+1,m+2,\dots}{
         
         
         \quad $||\textbf{L}^*||^1 = \min_{i \leq n}\{||\textbf{L}(\textbf{p}_i)||^1\}$, 
         \quad $\textbf{p}^{*} = \argmin_{\textbf{p}_{i \leq n}} ||\textbf{L}(\textbf{p}_i)||^1$  
         
         Select new hyperparameters $\vec{p}_{n+1} = \argmax EI_n(\textbf{p}_{n+1})$ using \\
         \quad $\mu = \boldsymbol{\mu}_{sum}(\textbf{p}_{n+1}), \sigma = \boldsymbol{\sigma}_{sum}(\textbf{p}_{n+1}), \textbf{L}(\textbf{p}_{n}^{*}) = ||\textbf{L}^*||^1$ in Eq. \ref{eq:EI}
         
         Train network $\mathcal{N}$ with hyperparameters $\vec{p}_{n+1}$ (using SGD) to obtain its parameters $\Theta_{n+1}$,
         loss $\textbf{L}_{n+1}$ and validation set performance $V_{n+1}$
         
         Augment data $\mathcal{D}_{n+1}$ = ${\mathcal{D}_n \cup (\textbf{p}_{n+1},\textbf{L}_{n+1},V_{n+1},\Theta_{n+1})}$
         
         Update surrogate model $\mathcal{M}(\boldsymbol{\mu},\boldsymbol{\sigma})$ using $\mathcal{D}_{n+1}(\textbf{p},\textbf{L})$

         }
         
         Choose best $\textbf{p}^\#$ and $\Theta^\#$ corresponding to best $V_i$ or minimum $\vec{L}_i$
         
        \KwRet{$\textbf{p}^\#,\Theta^\#$}
    }
    \caption{Bayesian Optimization for dCMF}
    \label{algo:bo}
\end{algorithm}

A common choice for the acquisition function is {\it Expected Improvement (EI)} \citep{jones2001taxonomy}, that has a closed form for GP, does not require its own tuning parameter and has been shown to perform well in minimization settings \citep{snoek2012practical}.
EI is the expectation that $\textbf{p}_{n+1}$ will improve $\textbf{L}$ (negatively, as we would like to minimize a loss) over $\textbf{p}_{n}^{*}$ which is 
the best observation from $n$ steps of BO so far, 
i.e. $\textbf{p}_{n}^{*} = \argmin_{\textbf{p}_{i \leq n}} \textbf{L}(\textbf{p}_i)$, and
$EI_n(\textbf{p}_{n+1}) = E_n[\max{\{(\textbf{L}(\textbf{p}_{n}^{*}) - \textbf{L}(\textbf{p}_{n+1})), 0\}}]$,
where the expectation $E_n$ is under the posterior distribution given evaluations of 
\textbf{L} at $\textbf{p}_1,\dots,\textbf{p}_n$. The next value is chosen by $\textbf{p}_{n+1} = \argmax EI_n(\textbf{p}_{n+1})$.
For a GP as $\mathcal{M}$, with predictive variance $\sigma(\textbf{p}_{n+1};\mathcal{D}_n,\mathcal{M})$ and predictive mean $\mu(\textbf{p}_{n+1};\mathcal{D}_n,\mathcal{M})$: 
\begin{equation}
\label{eq:EI}
EI_n(\textbf{p}_{n+1}) = \sigma [\gamma(\textbf{p}_{n+1}) \Phi(\gamma(\textbf{p}_{n+1})) + \phi(\gamma(\textbf{p}_{n+1}))]
\end{equation}
where $\gamma(\textbf{p}_{n+1}) = (\textbf{L}(\textbf{p}_{n}^{*}) -\mu)/\sigma$, and $\Phi$ and $\phi$ denote the CDF and PDF of the standard normal distribution respectively.

The extension of GP to vector-valued functions is through Multi-Task Gaussian Processes (MTGP), that can model outputs of multiple correlated tasks \citep{bonilla2008multi}.
\cite{swersky2013multi} demonstrate the advantages of MTGP as a surrogate model in several tasks with multiple dependent loss functions.

\noindent
{\bf MTBO for dCMF.}
A straightforward approach to solve the hyperparameter tuning problem for dCMF (problem (\ref{hyp})) is to use BO with GP as a surrogate model. 
However, we find that using MTGP within dCMF
shows better performance.
In dCMF, autoencoder training and matrix reconstruction tasks entail minimizing the sum of losses:
$\textbf{L} = \mathcal{L}_E + \mathcal{L}_R = \sum_{e=1}^E l_E^e + \sum_{m=1}^M l_R^m$.
Considering each of these as separate tasks, we have $E+M$ correlated tasks. We use MTGP as a surrogate model for BO, with the kernel specified through the intrinsic corregionalization model \citep{coburn2000geostatistics}:
$\mathcal{K}((p,t),(p',t')) = \mathcal{K}_t(t,t') \otimes \mathcal{K}_p(p,p')$
where $\otimes$ denotes the Kronecker product, $\mathcal{K}_p$ is a kernel measuring the similarity between the hyperparameters $\textbf{p}$ and $\mathcal{K}_t$ is the kernel measuring the similarities between the tasks. To ensure positive semidefiniteness of $\mathcal{K}_t$, it is parameterized through a Cholesky decomposition \citep{bonilla2008multi}: $\mathcal{K}_t = GG^T$, where $G$ is lower triangular. 
To model the dependencies between all the tasks, we initialize $\mathcal{K}_t$ as a unit matrix. 
Note that for each step in BO, MTGP yields an $(E+M)$-dimensional output. 

The EI acquisition function does not directly generalize to the multi-task case. So, \cite{swersky2013multi} use a heuristic approach, where a GP prior
is used for the average output of the tasks, and the the average predictive mean and predictive variance of multiple tasks
are used to select the next candidate. Instead, we use the sum of the predictive mean and variance (denoted by $\boldsymbol{\mu}_{sum}$ and $\boldsymbol{\sigma}_{sum}$ respectively) of each task since our final objective is to optimize the sum of losses. 
The output of the MTGP surrogate model $\mathcal{M}(\boldsymbol{\mu},\boldsymbol{\sigma})$ is scalarized (a common approach for multi-objective functions, e.g., in \citep{knowles2006parego}), by using the 1-norm. 
Denote the best value of the scalarized output by $||\textbf{L}^*||^1$.
Then our EI-based criterion can be computed using $\mu = \boldsymbol{\mu}_{sum}, \sigma = \boldsymbol{\sigma}_{sum}, \textbf{L}(\textbf{p}_{n}^{*}) = ||\textbf{L}^*||^1$ in equation \ref{eq:EI}.
Essentially, this heuristic chooses the next hyperparameter from regions where the tasks show high total predictive variance (exploration) or high total predictive mean (exploitation). 
We empirically evaluate this heuristic as the acquisition function with MTGP as the surrogate model and found it to be more effective than GP--based BO and random search for dCMF (see appendix \ref{ap:hyp}).
Algorithm \ref{algo:bo} shows the complete Bayesian Optimization strategy using MTGP as the surrogate model and our acquisition function heuristic.
The final hyperparameter set may be chosen based on the loss function or validation set performance.

\noindent
\subsubsection*{Complete dCMF Algorithm}

\begin{algorithm}[t]
    \caption{Deep Collective Matrix Factorization}
    \label{algo:dcmf}
    \SetKwInOut{Input}{inputs}
    \SetKwInOut{Output}{outputs}
    \SetKwProg{dCMF}{dCMF}{}{}
    \dCMF{$(G,\mathcal{X})$}{

        \Input{Entity-matrix relationship graph $G(V_E,V_M,D)$, \\ 
        Input matrices $\mathcal{X} = {X^{(1)},...,X^{(M)}}$}

        \Output{Entity representations $\mathcal{U}=U^{(1)},...,U^{(E)}$, \\ 
        Matrix reconstructions $\mathcal{X}^\prime = {X^{{(1)}^{\prime}},...,X^{{(M)}^{\prime}}}$}

        \tcp{Input Transformation}
        \ForEach{\textit{entity } $e_i \in V_E$ }{%
             $X_{list}$ = [$X^{(m)}$ \textbf{if} $(e_i,X^{(m)}) \in D$] \\
             Construct \textit{concatenated-matrix} $C^{(i)}$ = \textbf{concat}($X_{list}$)
        }

        \tcp{Network Construction}
        \ForEach{\textit{entity} $e_i \in V_E$ }{
             Construct $\textit{Autoencoder } \mathcal{A}^{(i)}$ 
        }
        Construct network $\mathcal{N}$ with $\mathcal{A}^{(i)}$ and collective loss $\textbf{L} = \mathcal{L}_E + \mathcal{L}_R $. 
        
        
        \tcp{Training and Hyperparameter Tuning} 
        \tcp{Run algorithm \ref{algo:bo} using network $\mathcal{N}$ and loss $\textbf{L}$}
        \tcp{Obtain best performing parameters $\Theta^\#$ and hyperparameters $\textbf{p}^\#$}
        $\textbf{p}^\#,\Theta^{\#}$ = \textbf{BO}($\mathcal{N}$,\textbf{L})

        \tcp{Entity representation generation}
        \ForEach{\textit{entity} $e_i$ in $V_E$ }{%
             $U^{(i)} = g^{(i)}_{{\textbf{p}^{\#},\theta^{\#}}}(C^{(i)})$ 
        }
        
        \tcp{Matrix reconstruction}
        \ForEach{\textit{matrix} $X^{(m)}$ in $V_M$ }{%
            $X^{(m)\prime}$ = $U^{(r_m)} \cdot U^{(c_m)^T}$
        }
        \KwRet{$\mathcal{U},\mathcal{X}^\prime$}
    }
\end{algorithm}

Algorithm \ref{algo:dcmf} shows the complete dCMF algorithm. Unless mentioned otherwise, we use MTGP as the surrogate model with the acquisition function described above for hyperparameter tuning.
We use stochastic gradient descent (SGD)
for training.\\
{\bf Loss functions.}
The loss functions $\mathcal{L}_R$ and $\mathcal{L}_E$ measure the model's average performance in reconstructing \textit{all} the entries of the input $X$ and concatenated $C$ matrices respectively. The error metric for $\mathcal{L}_R^{(m)}$ depends on the data type of $X^{(m)}$, e.g., root mean squared error for real values and cross entropy  for binary or categorical values. The choice of error metric for $\mathcal{L}_E$ is not straightforward if the concatenated matrix contains multiple data types. One way to address this is to transform the matrices $[X]_{e}$ associated with entity $e$, such that $C^{(e)}$ is of single data type (e.g. by scaling or PCA). We could also use multi-modal autoencoder architectures \citep{ngiam2011multimodal} designed to learn shared representations from multiple views of potentially different data types. 
In our experiments, we use root mean squared loss (for both $\mathcal{L}_R$ and $\mathcal{L}_E$). The root mean squared loss is more sensitive to larger errors and outliers as desired in the applications we present.
\\
{\bf Matrix Completion.}
Reconstruction of the matrices is obtained by multiplying the latent representations learnt for the corresponding row and column entities. 
Note that such a reconstruction yields real numbers that can be ordered and can be interpreted as scores for prediction or ranking tasks.

\noindent
{\bf Time Complexity.}
The training time complexity of dCMF is dominated by the autoencoder with largest input $C$ of dimension, say, $m \times d$ and is $\mathcal{O}(mdr)$ where $r$ is the number of neurons in the first layer. 
For BO, the time complexity is dominated by the matrix inversion step at each step for updating the MTGP model.
For $t = E+M$ tasks and $n$ steps in BO, the time complexity is $\mathcal{O}(n(t^3n^3 + mdr))$. 
For matrix completion, for a given matrix $X$ = $U^{(r)} \cdot U^{{(c)}^T}$, where $U^{(r)}$ and $U^{(c)}$ are inferred latent factors of dimensions $l \times K$ and $j \times K$, the time complexity is $\mathcal{O}(lKj)$, where $K$ is the assumed low rank.

\section{Experiments}
We first evaluate the performance of dCMF on various settings of sparsity level, size and shape of matrices, using synthetic data, to validate that dCMF addresses the optimization-related challenges discussed earlier.
We then evaluate the performance of dCMF on real-world benchmark datasets for two matrix completion tasks: movie recommendation and prediction of gene-disease association.
The source code for dCMF and data for all our experiments are available on our public repository.\footnote{\url{https://bitbucket.org/cdal/dcmf}}

\vspace{-1em}
\subsection{Effects of sparsity, size and shape}

We simulated datasets with 4 entities and 3 views based on the recommendation setup (fig. \ref{eg}(b)).
We generated $U^{(e_1)}$, $U^{(e_2)}$, $U^{(e_3)}$, $U^{(e_4)}$ with $K$=100 and the desired dimensions (mentioned below), with values sampled from a uniform distribution ranging between 0 and 1. We constructed the views $X^{(1)}_{m \times n}$, $X^{(2)}_{m \times u}$ and $X^{(3)}_{v \times n}$ using the corresponding factors, where subscripts indicate dimensions. 
To impart sparsity in a matrix $X^{(m)}$, random entries of the corresponding row/column entity factors $U^{(r_m)}$ or $U^{(c_m)}$ were set to zero until the desired level of sparsity was obtained. 
We use the Root Mean-Squared Error (RMSE) in predicting the central matrix $R = X^{(1)}$ as the performance measure, 
$
RMSE = \sqrt{\frac{1}{|T|}\sum_{R_{ij}\in T}({R_{ij} - R^\prime_{ij}})}
$,
where $R_{ij}$ is the ground truth and $R^\prime_{ij}$ the corresponding prediction. $T$ denotes the test set. 
In all experiments we perform 5-fold cross validation over the non-zero entries of the central matrix $R$. 

\begin{figure*}[!h] 
\centering
        \subfloat[]{
        \centering
                \fontsize{12pt}{12pt}\selectfont
                \includegraphics[width=0.2875\textwidth]{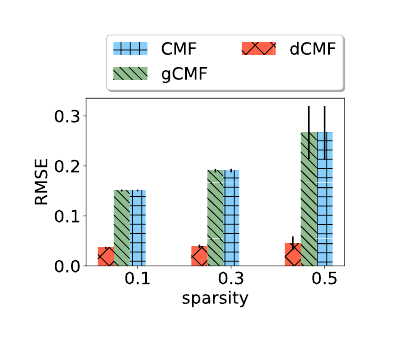}
                \label{fig:simulation_results:sparsity}       
        }
        \subfloat[]{
        \centering
                \fontsize{12pt}{12pt}\selectfont
                \includegraphics[width=0.3\textwidth]{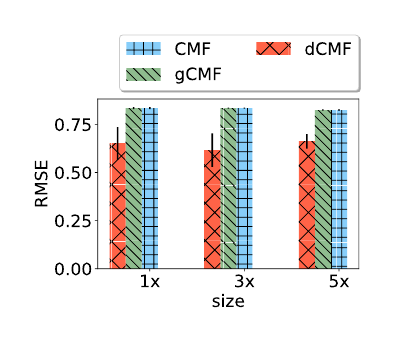}
                \label{fig:simulation_results:size}           
        }
        \subfloat[]{
        \centering
                \fontsize{12pt}{12pt}\selectfont
                \includegraphics[width=0.3\textwidth]{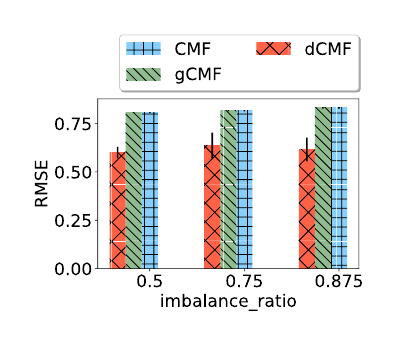}
                \label{fig:simulation_results:shape}
        }
        \caption{Impact of (a) Sparsity (b) Entity size and (c) View shape on performance of dCMF and CMF using synthetic datasets.}\label{fig:simulation_results}
\end{figure*}

\noindent
{\bf Sparsity.}
Consider sparsity level as the proportion of zero entries in the central matrix $X^{(1)}$.
To illustrate how sparsity impacts the performance of dCMF, 
we simulated 3 artificial datasets with same dimensions ($m=1000,\,\,n=2000,\,\,u=200,\,\, v=400$) and increasing sparsity levels 0.3, 0.5 and 0.7.
25,000 non-zero entries randomly chosen from the test fold was used as the test set. This is to ensure that we measure RMSE over the same test set
(since varying sparsity levels varies the number of non-zeros entries and thereby  the test fold size).
CMF, gCMF and dCMF were used  (with $K$ set to 100) to predict the entries in the test set. No input transformation was performed for these experiments.
It can be seen from fig. \ref{fig:simulation_results}\subref{fig:simulation_results:sparsity} that increase in sparsity results in increased RMSE in CMF, gCMF and dCMF, with dCMF consistently outperforming CMF and gCMF. 

\noindent
{\bf Size and Shape.}
We simulated the first dataset with dimensions $m=400,\,\,n=800,\,\,u=80,\,\, v=160$. 
Then we created two other datasets that are 3 and 5 times the size of the first one. Fig. \ref{fig:simulation_results}\subref{fig:simulation_results:size} shows the performance of CMF, gCMF and dCMF.
We define \textit{imbalance-ratio} of a view with shape $m \times n$ as $ \left (1 - \frac{\min(m,n)}{\max(m,n)} \right )$. 
Thus the \textit{imbalance-ratio} is 0 if $m=n$ and increases otherwise.
We created 3 datasets with $n=2000, u=200, v=400$ and increasing \textit{imbalance-ratios} by varying $m$:  0.5 ($m=1000$), 0.75($m=500$) and 0.875 ($m=250$). Fig. \ref{fig:simulation_results}\subref{fig:simulation_results:shape} shows the performance of all three methods (with $K$ set to 100) on these datasets. 
We find that BO is able to effectively select hyperparameters for different sizes and imbalance ratios and dCMF consistently outperforms CMF and gCMF in all the settings.

\subsection{Case Study: Hybrid Recommender Systems}
Among the large number of recommendation algorithms developed, arguably, the most well known are Collaborative Filtering (CF) methods 
that factorize the historical user-item rating matrix
to obtain latent user representations.
Content-based methods use item descriptions or user profiles
to recommend items that are similar to items found in a user's history
(e.g. \citep{pazzani1997learning}).
CF has been more successful than content-based methods 
but suffers from two problems:
(1) real world rating matrices are large and sparse which impacts the latent factors learnt and deteriorates recommendation performance, (2) they cannot be used to recommend items to a user with no previous ratings,
known as the {\it cold-start} problem.
Hybrid methods combine the strengths of both these methods by incorporating user and item information as {\it side information} within CF.

Deep learning models have been successful in obtaining good representations 
in recommender systems.
Among the earliest models, is Collaborative Deep Learning (CDL) \citep{wang2015collaborative}, that jointly performs deep representation learning for the content side information and collaborative filtering for the rating matrix.
To obtain these representations stacked denoising autoencoders \citep{vincent2010stacked} are used in a Bayesian formulation. 
A more scalable and efficient architecture that combined CF with marginalized denoising autoencoders \citep{chen2012marginalized} was used in Deep Collaborative Filtering (DCF) \citep{li2015deep}.
CDL has also been recently extended to model multimedia side information in a more robust manner in Collaborative Variational AutoEncoder (CVAE) \citep{li2017collaborative}.
Note that CDL and CVAE model only the rating and content matrices and not user side information.
The use of additional side information was leveraged in the additional stacked denoising autoencoder (aSDAE), that was designed to integrate side information into the latent factors efficiently. 
Using a combination of aSDAE and matrix factorization, was shown to outperform CDL and DCF \citep{dong2017hybrid}.
Another variant of the stacked denoising autoencoder was used with Convolutional Neural Networks to generate user and item latent features respectively and combined in probabilistic model called Probabilistic HybriD model (PHD) \citep{liu2017phd} that was shown to outperform aSDAE.


\noindent
{\bf Recommendation with dCMF.} \label{subsection: reco_deep_cmf}
A typical recommendation setting with side information 
contains 4 entities: users, items, user-features and item-features (of dimensions $m, n, u, v$ respectively) and 3 matrices as shown in fig. \ref{eg}(b) and described in table \ref{tab:reco}.
Matrix $X^{(1)}$ is usually very sparse due to unknown ratings and the recommendation task is to complete this matrix to obtain future movie recommendations for users.

\noindent
{\bf Prediction.}
Prediction is directly obtained through matrix completion by multiplying the latent representations learnt for row and column entities of the rating matrix as $X^{{(1)^\prime}}$ = $U^{(e_1)} \cdot U^{{(e_2)}^T}$. 
Similar to other CMF-based methods, dCMF can be used to address the cold-start problem. 
Note that $U^{(e_1)}$ will not contain an entry for a first time user and so, $X^{{(1)}^\prime}$ 
will also not contain recommendations for this user. To overcome this cold-start problem, we use 
${X^{(2)}}_{m \times u}$ = $U^{(e_1)}_{m \times K} \cdot U_{u \times K}^{(e_3)^T} \implies U^{(e_1)} = {X^{(2)}}_{m \times u} \cdot ({U_{u \times K}^{(e_3)^{T}})^{-1}}$, which can be used to estimate an unknown user's latent factor. 
For a single user's feature vector $h_{1 \times u}$, this yields the recommendation: 
$X^{(1)^\prime}_{1 \times n} = (h_{1 \times u} \cdot ({U_{u \times K}^{(e_3)^{T}}})^{-1}) U^{(e_2)^T}_{n \times K}$.

\noindent
{\bf Data.}
We use two large benchmark datasets: (1) MovieLens-100K
and (2) MovieLens-1M.
The ratings are between 1 and 5 (star ratings). Both the datasets contain user demographic information (age, gender, occupation, zip) and movie metadata (title, genre). We constructed the rating matrix $X^{(1)}$ by binarizing the ratings (to predict user-movie associations) 
and bag-of-words feature matrices $X^{(2)}$ and $X^{(3)}$, as described in \citep{dong2017hybrid}. 
Matrix statistics after feature processing are shown in table \ref{tab:reco}.

\begin{table}[!h]
    \centering
    \small
    \begin{tabular}{|c|c|c|c|c|c|c|}
    \hline
        \multicolumn{3}{|c|}{Datasets:}  & \multicolumn{2}{|c|}{MovieLens-100K} & \multicolumn{2}{|c|}{MovieLens-1M}\\ \hline
        Matrix & Row Entity & Col Entity & Row Dim & Col Dim &  Row Dim & Col Dim\\ \hline
        $X^{(2)}$ & User & User Features & 943 & 823& 6040 & 3467\\ \hline
        $X^{(1)}$ & User & Movies & 943& 1682 & 6040 & 3706 \\ \hline
        $X^{(3)}$ & Movie Features & Movies & 2374 & 1682 & 4296 & 3706 \\ \hline
    \end{tabular}
    \caption{Recommendation Dataset Statistics (Col: Column, Dim: Dimension)}
    \label{tab:reco}
\end{table}

\noindent
{\bf Baselines.}
We compare our performance with state-of-the-art hybrid recommendation algorithms that use both the row (user) and column (movie) features of the central matrix (rating). 
Our main baseline is aSDAE \citep{dong2017hybrid}, a hybrid recommender model that was shown to outperform DCF, CDL, CMF and PMF.
Note that aSDAE can be used for recommendation with side-information but cannot be used for augmented multi-view learning.
In addition we also compare with PHD \citep{liu2017phd}, DCF \citep{li2015deep},  and IMC \citep{natarajan2014inductive}, that can use side information for recommendations.
We also compare our performance with that of CMF and gCMF that use collective matrix factorization.

\noindent
{\bf Evaluation metric.}
Since these ratings are implicit and we do not evaluate ranking, we use Recall@N (averaged over all users) as our evaluation metric. \cite{dong2017hybrid} use Recall@N for the same reason.
Let $r_i$ be the row corresponding to the predictions for user $i$ in the predicted rating matrix $X^{(1)^\prime}$, $S_i^N$ be the set of top $N$ predictions from the sorted $r_i$ and $S_i^T$ be the test set for the user $i$. 
Recall@N = $\frac{|S_i^N \cap S_i^T|}{|S_i^T|}$.
Following  \cite{dong2017hybrid}, for each dataset, we measure the average performance over 5 runs. In each run, 95\% of the ratings are randomly selected for training and the remaining 5\% for the test set. 
P-values are computed using the Friedman test \citep{demvsar2006statistical}.

\begin{figure*}[!h] 
\centering
        \subfloat[MovieLens-100K]{
        \centering
                \fontsize{6pt}{6pt}\selectfont
                \includegraphics[width=0.4\textwidth]{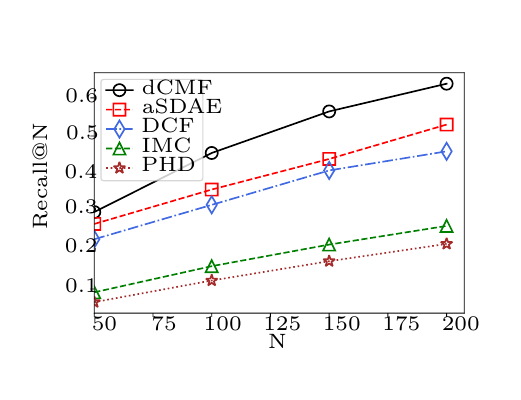}
                \label{fig:real_data_results:ml100k:graph1}        
        }
        \subfloat[MovieLens-100K]{
        \centering
                \fontsize{6pt}{6pt}\selectfont
                \includegraphics[width=0.4\textwidth]{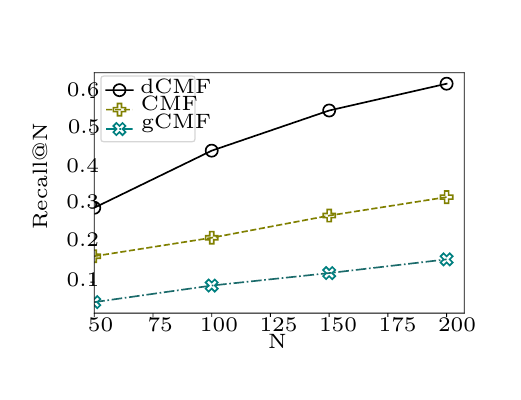}
                \label{fig:real_data_results:ml100k:graph2}             
        } \\
        \subfloat[MovieLens-1M]{
        \centering
                \fontsize{6pt}{6pt}\selectfont
                \includegraphics[width=0.4\textwidth]{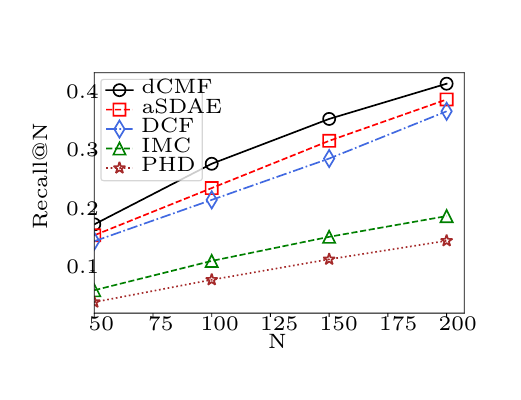}
                \label{fig:real_data_results:ml1m:graph1}
        }
        \subfloat[MovieLens-1M]{
        \centering
                \fontsize{6pt}{6pt}\selectfont
                \includegraphics[width=0.4\textwidth]{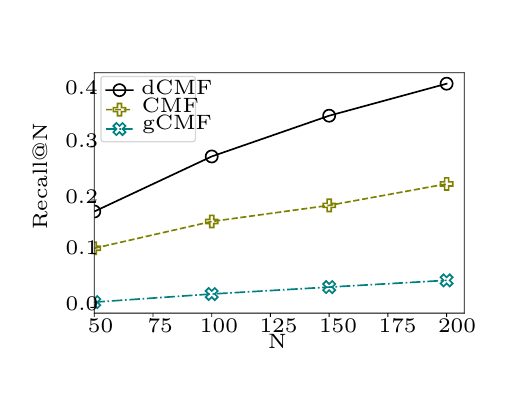}
                \label{fig:real_data_results:ml1m:graph2}
        }
        \caption{Performance of dCMF and baseline methods ((left) recommendation algorithms, (right) CMF-based algorithms) on two benchmark datasets.
        }
        \label{fig:real_data_results}
\end{figure*}

\noindent
{\bf Results.}
Fig. \ref{fig:real_data_results} (\subref{fig:real_data_results:ml100k:graph1}
 and \subref{fig:real_data_results:ml100k:graph2})
  show the performance of all the methods on the MovieLens-100K dataset. 
Fig. \ref{fig:real_data_results}(\subref{fig:real_data_results:ml1m:graph1} and 
\subref{fig:real_data_results:ml1m:graph2}) show the performance results on MovieLens-1M dataset
(recommendation and CMF baselines shown separately). 
In both the datasets, we observe that dCMF significantly outperforms all the baseline hybrid recommendation methods
aSDAE, DCF, IMC and PHD 
(p-value 0.003)
, as well as previous CMF methods, CMF and gCMF 
(p-value 0.018).

\subsection{Case Study: Gene-disease association prediction}

Identifying the genes associated with diseases is an important problem in biomedical sciences. 
Knowledge of such associations not only improve our understanding genomic interactions but also facilitate the design of treatment strategies.
As a result, there has been active research in this area with many experimental methods to determine such associations such as 
genome-wide association studies (GWAS) \citep{frayling2007genome} and RNA 
interference screens \citep{boutros2008art}.
However experimental methods are expensive, time-consuming and may be specific to certain classes of diseases \citep{piro2012computational}.
As a result various computational approaches have been developed to aid the discovery of such associations, such as knowledge-based methods (e.g., \citep{zhou2016knowledge}) and methods based on text mining (e.g., \citep{kolker2015finding}), crowdsourcing  (e.g., \citep{loguercio2013dizeez}) and networks (e.g., \citep{singh2013prediction,zeng2017prediction}).
Comprehensive surveys of these methods can be found in \citep{piro2012computational,
opap2017recent,seyyedrazzagi2017disease}.

Tremendous heterogeneity can be found in biological data -- comprising measurements from diverse aspects of our complex biological systems -- that are used to infer gene-disease associations. When the evidence for an association can be found through multiple independent sources, it is more likely to be true; indeed, methods that can leverage the heterogeneity have been reported to have superior performance \citep{pers2011meta}.
Hence, many heterogeneous network based methods have been developed for predicting gene-disease association.
For example, HSSVM \citep{zeng2017prediction} and CATAPULT \citep{singh2013prediction}, both can integrate different biological networks (like protein-protein interactions, disease-disease similarities) and also relevant data from other species.
The main limitation of such methods is that they cannot be used for genes or diseases with no known associations (similar to the cold-start problem in recommendation).

A matrix completion based approach, Inductive Matrix Completion (IMC), was proposed by \cite{natarajan2014inductive} where the problem is modeled as a recommendation problem. 
Genes and diseases are analogous to users and movies respectively, the rating matrix is analogous to the gene-disease association matrix which is also partially observed and sparse.
Similar to hybrid recommender systems, side information as features for genes and diseases can be used from various data sources, to improve the predictive accuracy of the model. 
Their method is found to significantly outperform previous best methods that cannot integrate multiple data sources. Further, their method can also predict associations for genes or diseases with no previously known associations.

However IMC is limited to using features of genes or diseases, i.e., only data that can be transformed into the matrices described in fig. \ref{eg}(b): gene features ($X^{(2)}$), disease features ($X^{(3)}$) and gene-disease associations ($X^{(1)}$). Any other auxiliary source of information that may be pertinent to discovering gene-disease association cannot be incorporated.
Since CMF-based methods can obtain latent representations from arbitrary collection of matrices, such auxiliary information can be modeled.
We show that for gene-disease prediction, such sources, indeed improve the performance of predicting gene-disease association,
in an augmented multi-view setting (fig. \ref{fig:dcmf_overview}\subref{fig:dcmf_overview:b}).


\noindent
{\bf Data.}
We use four publicly available biomedical data sources: 
\begin{enumerate}
\item 
\textit{DisGeNET} (\cite{pinero2016disgenet}) is a database of known gene-disease associations, collected from expert curated repositories, GWAS catalogues, animal models and scientific literature.
\item 
\textit{The Cancer Genome Atlas (TCGA)} (\cite{weinstein2013cancer}) 
contains genomic and clinical data of 33 different cancers and over 10,000 patients.
\item 
\textit{Humannet} (\cite{lee2011prioritizing}) is a functional gene network of human genes obtained by integration of 21 types of `omics' data sources. Each edge in HumanNet is associated with the probability of a true functional linkage between two genes.
\item 
\textit{UMLS Metathesaurus} (\cite{schuyler1993umls}) is a 
large database of biomedical concepts and their relationships.
\end{enumerate}

We only consider the expert curated gene-disease associations from DisGeNET for our dataset construction, since these are the most reliable.
We also restrict our data to a single cancer (Breast Cancer) in TCGA. With these restrictions, there were 11939 genes that were present in all three databases: DisGeNET, TCGA and HumanNet, with 1093 and 11809 associated patients and diseases respectively. 
We chose a random subset of 2000 genes and associated diseases (968) and all the patients (1093).
For these genes and diseases, the gene-disease association matrix $X^{(1)}$ was constructed by using all known associations from DisGeNET: there were 69850 associations, resulting in sparsity level of 96.5\%.
To construct $X^{(2)}$ we used RNA-Seq Expresssion data from TCGA, where a single sample per patient was chosen.
TCGA also contains 115 demographic and clinical features for these patients. We chose a subset of 8 numeric and 21 categorical features as listed in Table \ref{table:patient_features} in the Appendix \ref{ap:gda}, with the less than 50\% missing values. We then transformed the categorical features to their one-hot encodings, normalized the numeric features and obtained a total of 86 patient features.
Gene-gene and disease-disease graphs for the selected genes and diseases were obtained from HumanNet and UMLS respectively.
Similar to preprocessing done by \cite{natarajan2014inductive}, we use principal components of the adjacency matrices of these graphs as features to obtain matrices $X^{(3)}, X^{(4)}$.
Note that this dataset forms the augmented multi-view setup shown in fig. \ref{fig:dcmf_overview}. Table \ref{tab:dataset} shows the entity type for each matrix and the matrix dimensions.

\begin{table}[h]
    \centering
    \small
    \begin{tabular}{|c|c|c|c|c|}
         \hline
         Matrix & Row Entity & Col Entity & Row Dim & Col Dim  \\ \hline
         $X^{(1)}$ & Gene & Disease & 2000 & 968 \\ \hline
         $X^{(2)}$ & Gene & Patient & 2000 & 1093 \\ \hline
         $X^{(3)}$ & Gene & Gene Features & 2000 & 1000 \\ \hline
         $X^{(4)}$ & Disease Features & Disease & 500 & 968 \\ \hline
         $X^{(5)}$ & Patient Features & Patient & 86 & 1093 \\ \hline
    \end{tabular}
    \caption{\small Gene-Disease Association: Dataset Statistics (Col: Column, Dim: Dimension)}
    \label{tab:dataset}
\end{table} 

\noindent
{\bf Baselines.}
IMC has been found to outperform heterogeneous network based methods like CATAPULT \citep{singh2013prediction}. In a recent work, another heterogeneous network based method HSSVM \citep{zeng2017prediction} was proposed but it could not outperform CATAPULT. So, we use IMC as the main baseline for predicting gene-disease associations. Note that IMC cannot utilize information from $X^{(5)}$ (patient-patient\_features). The other baselines are CMF and gCMF that can model all the views.

\noindent
{\bf Evaluation Metric.} As discussed in \citep{natarajan2014inductive}, an appropriate metric for this task is \textit{probability@N}. 
For each disease in the test set, the genes are ranked by the score predicted by each method.
The cumulative distribution of the ranks, \textit{probability@N}, is the probability that the rank at which a hidden gene-disease pair is retrieved is less than a threshold N.
We created 5 folds of the gene-disease association matrix entries (using only known associations) for cross validation.
We report the \textit{probability@N} averaged over the 5 folds for N ranging from 1 to 100.  
P-values are computed using the Friedman test \citep{demvsar2006statistical}.
\begin{figure}[h!] 
        \centering
                \def\svgwidth{0.55\textwidth}       
                \begingroup\makeatletter\def\f@size{10}\check@mathfonts 
                \def\maketag@@@#1{\hbox{\m@th\large\normalfont#1}}%
                \includegraphics[width=0.55\textwidth]{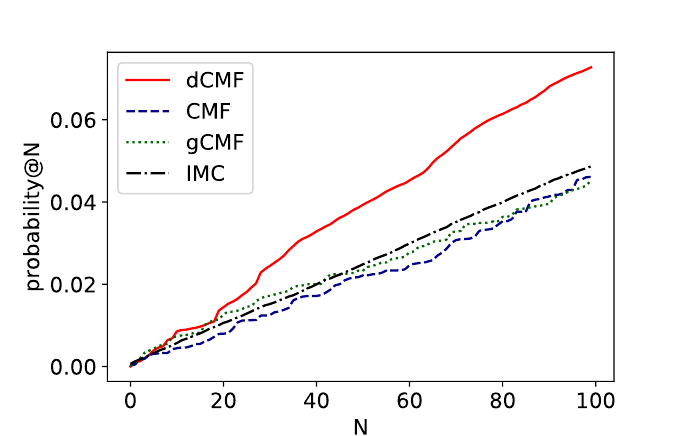}
                \endgroup
                \caption{Gene-Disease Association Prediction: Performance of dCMF and baselines.}
                \label{bo:aug}       
\end{figure}

\noindent
{\bf Results.}
Fig. \ref{bo:aug} shows the performance of dCMF, gCMF, CMF and IMC on our dataset for different values of N. The performance of IMC, CMF and gCMF are comparable with IMC doing marginally better than CMF and gCMF. While dCMF is comparable to IMC below N=10,  dCMF significantly outperforms all three baselines at all values of N above 10
(p-value $<0.0001$).

Although all three CMF-based methods can utilize the information in the matrix $X^{(5)}$, which IMC cannot, only dCMF can outperform IMC.
This suggests that by modeling non-linear interactions, dCMF is better than CMF and gCMF, at integrating heterogeneous data for predictive modeling.

\section{Conclusion}

We present dCMF, a neural architecture for CMF, that,
to our knowledge, is the first deep augmented multi-view learning technique.
dCMF effectively learns latent entity representations, shared across multiple matrices and models their non-linear interactions, that previous CMF methods cannot.
Our empirical results show that by modeling non-linear interactions, dCMF effectively integrates heterogeneous data sources and obtains shared representations for predictive modeling that are better than those of several state-of-the-art methods.

Learning dCMF model parameters involves optimizing both entity-specific autoencoder losses as well as matrix-specific reconstruction losses.
The latter induces a dependency between the latent representations, which necessitates principled hyperparameter tuning to scale our neural architecture to an arbitrary collection of matrices.
Through multi-task Bayesian optimization and an acquisition function that is adapted for dCMF, we effectively address these challenges.
Our experiments demonstrate that dCMF significantly outperforms previous CMF methods in both simulated and real datasets.
We demonstrate two applications of dCMF: movie recommendations and prediction of gene-disease associations. In both tasks, dCMF significantly outperforms state-of-the-art algorithms on three benchmark datasets.

This work can be extended in several ways.
To address the problem of mixed sparsity levels in the input matrices,
we could explore other architectural variants. 
E.g., architectures similar to that in \citep{ngiam2011multimodal}
could be used, that can also model view-specific noise and naturally handle different data types.
The effect of other types of autoencoders, such as variational autoencoders \citep{kingma2013auto}, could also be studied further.
Techniques to improve the scalability of training and hyperparameter tuning can be explored.
Finally negative transfer, that is known to affect CMF \citep{lan2016towards}, requires further investigation within the dCMF architecture.

\bibliography{biblio}
\bibliographystyle{spr-chicago}

\appendix

\section{Hyperparameters}\label{ap:hyp}



We briefly describe the hyperparameters that can be tuned in dCMF.
The model hyperparameters include the entity representation dimension $K$ and the fraction $f_k$, that decide the number of neurons/units in each layer and the number of layers adaptively based on the input dimension.
The optimization related hyperparameters to be searched include learning rate, weight-decay, batch size, maximum epochs and convergence threshold. Table \ref{tab:hyperparams} lists all the hyperparameters.

\begin{table}[!h]
\resizebox{\textwidth}{!}{%
\begin{tabular}{ll}
Learning Algorithm Parameters:                   & Model Parameters:                           \\
\hline
Learning rate                                    & Fraction $f_k$ (Number of layers; Number of neurons per layer) \\
Convergence Threshold                            & Encoding and decoding activation function choice              \\
Weight Decay                                     & Entity representation size $k$                                \\
Batch Size                                       &                                                               \\
Maximum Epochs                                   &                                                               \\
Pre-training Requirement (Convergence Threshold) &                                                              
\end{tabular}%
}
\caption{List of dCMF Hyperparameters}
\label{tab:hyperparams}
\end{table}

\subsection{Hyperparameter Optimization}

\subsubsection{Evaluation of Acquisition Function Heuristic}

We illustrate the effect of our acquisition function heuristic through an example.
We consider a setup with two outputs and a single input. The two tasks are defined by the functions $f_1(x)$ and $f_2(x)$ below, where $x$ is the single one dimensional input.

	$$f_1(x) = sin(x) + sin((10/3)*x)$$
	$$f_2(x) = 2 cos(x) + cos(2x)$$

\begin{figure*}[!t] 
\centering
        \subfloat[]{
        \centering
                \fontsize{10pt}{10pt}\selectfont
                \includegraphics[width=0.25\textwidth]{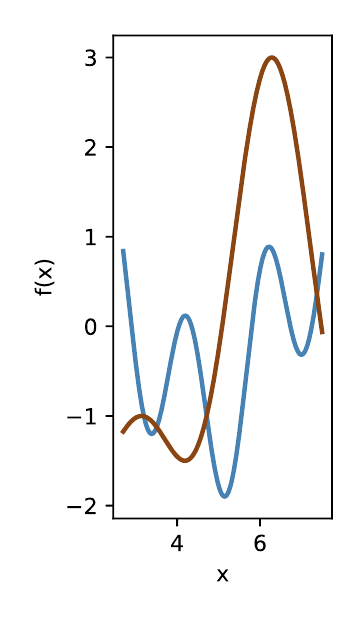}
                \label{bo:ei_all:a}       
        }
        \subfloat[]{
        \centering
                \fontsize{10pt}{10pt}\selectfont
                \includegraphics[width=0.225\textwidth]{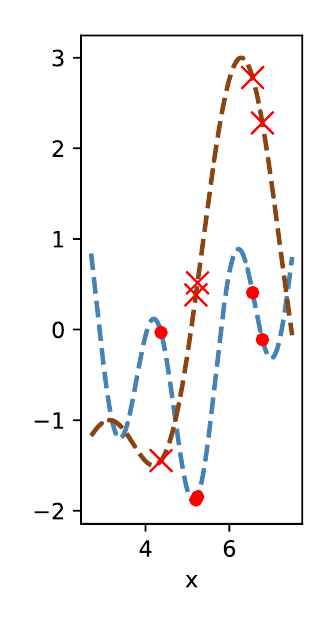}
                \label{bo:ei_all:b}           
        }
        \subfloat[]{
        \centering
                \fontsize{10pt}{10pt}\selectfont
                \includegraphics[width=0.445\textwidth]{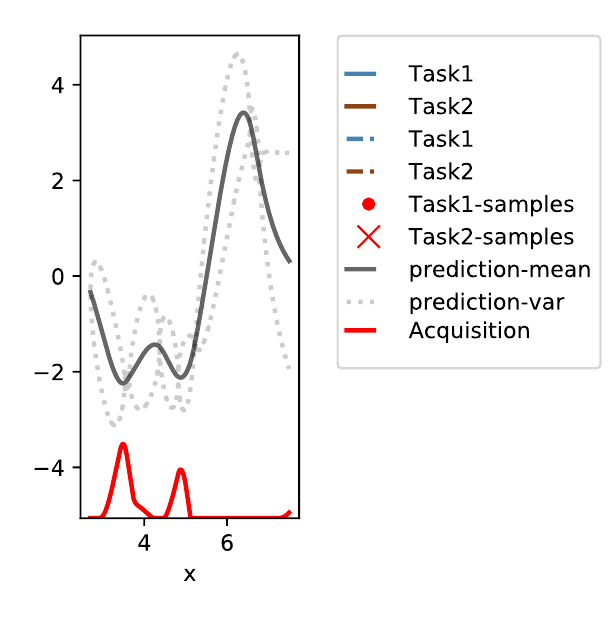}
                \label{bo:ei_all:c}
        }
        \caption{Illustration of Expected Improvement based heuristic}
        \label{bo:ei_all}
\end{figure*}

The two functions are shown in fig. \ref{bo:ei_all}(a) for the domain of $ x \in [2.5,7.5]$. As an initial design for BO we used $5$ randomly sampled $x$ values and the corresponding $f_1(x)$ and $f_2(x)$ to train the surrogate model (MTGP). We then performed $5$ BO steps and the corresponding samples selected based on our acquistion function are shown in fig. \ref{bo:ei_all}(b). We illustrate the acquisition function values against the MTGP predictive mean and variance's 1-norm in fig. \ref{bo:ei_all}(c). It can be seen that the acquistion function peaks correspond to high variance or high (negative) mean. 

\subsubsection{Evaluation of Surrogate Model}

\begin{figure}[!h]
    \centering
        \def\svgwidth{0.6\textwidth}       
        \begingroup\makeatletter\def\f@size{10}\check@mathfonts 
        \def\maketag@@@#1{\hbox{\m@th\large\normalfont#1}}%
        \includegraphics[width=0.6\textwidth]{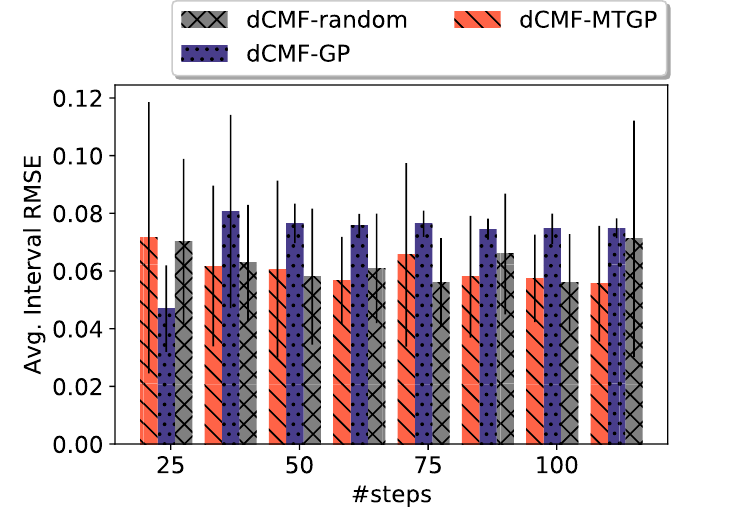}
        \endgroup
        \caption{Surrogate model selection for hyperparameter tuning}
                \label{bo:mtgp} 
\end{figure}

To evaluate our approach to hyperparameter selection, we constructed a synthetic dataset in the augmented multi-view setting shown in fig. \ref{fig:dcmf_overview}(b). The dataset consists of 6 entities $e_1,\ldots,e_6$ of dimensions 1000, 2000, 20, 150, 300 and 250 respectively. 
dCMF (Algorithm \ref{algo:dcmf}) was run with three different choices of algorithm \ref{algo:bo}: 
(1) BO with GP \citep{snoek2012practical} (denoted by dCMF-GP), 
(2) Random search \citep{bergstra2012random} (denoted by dCMF-random)
(3) BO with MTGP, as described above using our acquisition function heuristic (denoted by dCMF-MTGP).
We set $n = 200$ steps in Algorithm \ref{algo:bo}.  
At every step we reconstruct the matrix $X^{(1)}$ and compute $l_R(X^{(1)},X^{(1)\prime})$ on a held-out test set using RMSE. We use average cumulative RMSE computed in intervals of 25 steps as our evaluation criterion. Fig. \ref{bo:mtgp} shows that dCMF-MTGP has the lowest RMSE after 100 steps, while dCMF-GP initially has the lowest RMSE but is consistently higher after 50 steps. dCMF-random does not have a consistent performance. 

\subsection{dCMF Hyperparameter Settings for Case Studies} 

In this section we list the hyperparameter settings, found using MTBO (algorithm \ref{algo:bo}), that was used in our experiments.

\subsubsection{Hybrid Recommender System}

We set tanh as the activation function in all the encoding and decoding layers. Setting the activation function as tanh allows the range of the learnt factors $U$ to be between -1 and +1. This provides flexibility in reconstructing input matrices $X$ thereby lowering the matrix reconstruction loss $\mathcal{L}_R$. For the recommendation datasets we empirically found tanh to do better than ReLu.
During training, data was used in 2 batches. We did pretraining for MovieLens-100K and not for MovieLens-1M dataset as the pretraining did not improve the performance. 
In the side matrices, we perform Maximum Absolute Scaling in which we do not shift/center the data but translate each feature such that their maximal absolute value is 1.0.
With the manual settings described so far following are the best hyperparameters $\textbf{p}^*$ as found by BO in 200 steps:  $f_k=0.01, k=200$, learning rate $10^{-4}$ and convergence threshold $10^{-5}$(MovieLens-100K) \& $10^{-4}$(MovieLens-1M).

\subsubsection{Gene Disease Association Prediction} 

We set tanh as the  activation function in all the encoding and decoding layers. We did not do pretraining and data was used as a single batch during training. With these manual settings following are the best hyperparameters $\textbf{p}^*$ as found by BO in 200 steps: $f_k=0.6, k=100$, learning rate $0.0002$, and convergence threshold $0.0006$. 





\section{Dataset details for Gene Disease Association Prediction}\label{ap:gda} 

The list of patient features selected from the TCGA dataset for our case study on gene-disease prediction is shown in table \ref{table:patient_features}.

\begin{table}[!h]
\resizebox{\textwidth}{!}{%
\begin{tabular}{@{}ll@{}}
\toprule
CATEGORICAL                                                               & NUMERIC                                                                     \\ \midrule
\hline
American Joint Committee on Cancer Tumor Stage Code                       & Diagnosis Age                                                               \\
Neoplasm Disease Lymph Node Stage American Joint Committee on Cancer Code & Death from Initial Pathologic Diagnosis Date                                \\
American Joint Committee on Cancer Metastasis Stage Code                  & Positive Finding Lymph Node Hematoxylin and Eosin Staining Microscopy Count \\
Neoplasm Disease Stage American Joint Committee on Cancer Code            & Disease Free (Months)                                                       \\
New Neoplasm Event Post Initial Therapy Indicator                         & Lymph Node(s) Examined Number                                               \\
Metastatic tumor indicator                                                & Last Alive Less Initial Pathologic Diagnosis Date Calculated Day Value      \\
Overall Survival Status                                                   & HER2 ihc score                                                              \\
Disease Free Status                                                       & Overall Survival (Months)                                                   \\
Patient's Vital Status                                                    &                                                                             \\
ER Status By IHC                                                          &                                                                             \\
Prior Cancer Diagnosis Occurence                                          &                                                                             \\
Micromet detection by ihc                                                 &                                                                             \\
PR status by ihc                                                          &                                                                             \\
Person Neoplasm Status                                                    &                                                                             \\
Ethnicity Category                                                        &                                                                             \\
Tissue Retrospective Collection Indicator                                 &                                                                             \\
Disease Surgical Margin Status                                            &                                                                             \\
Sex                                                                       &                                                                             \\
Primary Lymph Node Presentation Assessment Ind-3                          &                                                                             \\
Neoadjuvant Therapy Type Administered Prior To Resection Text             &                                                                             \\
Tissue Prospective Collection Indicator                                   &                                                                             \\ \bottomrule
\end{tabular}%
}
\caption{List of patient features.}
\label{table:patient_features}
\end{table}

\section{Model Complexity}\label{ap:k} 
In this section we empirically investigate the following: Is the performance improvement due to dCMF solely because of larger number of model parameters?
In other words, if CMF or gCMF were to use larger number of free parameters, would their performance improve and be similar to that of dCMF?

We first analyze the number of free parameters in CMF, gCMF and dCMF:

\noindent
{\bf CMF.}
The number of parameters in CMF $\text{p}_\textsubscript{cmf} = \text{p}_\textsubscript{u}$, where $\text{p}_\textsubscript{u} = \sum_{e \in E}(|e|*K)$.
CMF extends the alternating projection method and uses a Newton-Raphson step in a gradient-descent based algorithm to estimate all the latent factors
\citep{singh2008relational}.

\noindent
{\bf gCMF.}
In gCMF a variational Bayesian solution is developed wherein additional parameters are present for the distributions assumed.
So, the number of parameters, $\text{p}_\textsubscript{gcmf}$ = $\text{p}_\textsubscript{u} + \text{p}_\textsubscript{g}$, $$\text{p}_\textsubscript{g} = |\{\tau^{(m)}\}_{m \in M}| + |\{\alpha_{e,k}\}_{e \in E, k \in (1 \dots K)}| + |\{\mu_{(e,m)}, \sigma^2_{(e,m)}\}_{e \in E, m \in M}| + |\{p_0, q_0, a_0, b_0\}|$$ 
where, $\alpha_{e,k}$: Gaussian likelihood precision for latent factors, $\tau^{(m)}$: precision for error terms, \{$p_0, q_0, a_0, b_0$\}: gamma prior parameters and $\mu_{(e,m)}$: mean and $\sigma^2_{(e,m)}$: variance of the bias terms. $|.|$ denotes set cardinality.

\noindent
{\bf dCMF.}
The number of parameters in dCMF,  $\text{p}_\textsubscript{dcmf} = \text{p}_\textsubscript{u} + \text{p}_\textsubscript{d}$, where $\text{p}_\textsubscript{d} = {\sum_{e \in E}|}\{\text{\textit{par}}(\mathcal{A}^{(e)})\}|$ and $\text{\textit{par}}(\mathcal{A}^{(e)})$ are the parameters associated with the autoencoder corresponding to entity $e$.

\noindent
Note that $\text{p}_\text{cmf} = \text{p}_\text{u}$, $\text{p}_\text{gcmf} \approx \mathcal{O}(\text{p}_\text{u})$ and $\text{p}_\text{dcmf} \approx \mathcal{O}(\text{p}_\text{u} + \text{p}_\text{d})$ and by varying $K$ we can control the model complexity. 

\begin{figure*}[t] 
\centering
        \subfloat[]{
        \centering
                \fontsize{12pt}{12pt}\selectfont
                \includegraphics[width=0.45\textwidth]{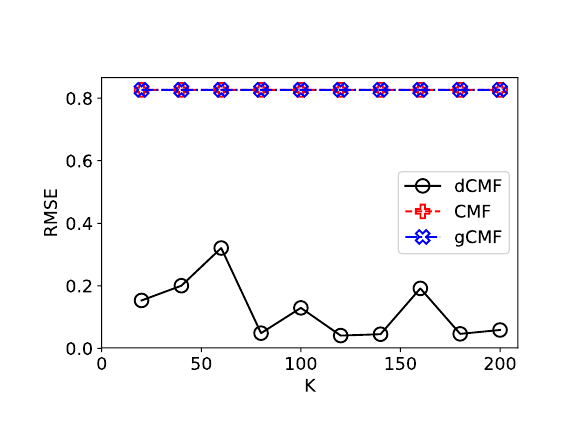}
        }
        \subfloat[]{
        \centering
                \fontsize{12pt}{12pt}\selectfont
                \includegraphics[width=0.48\textwidth]{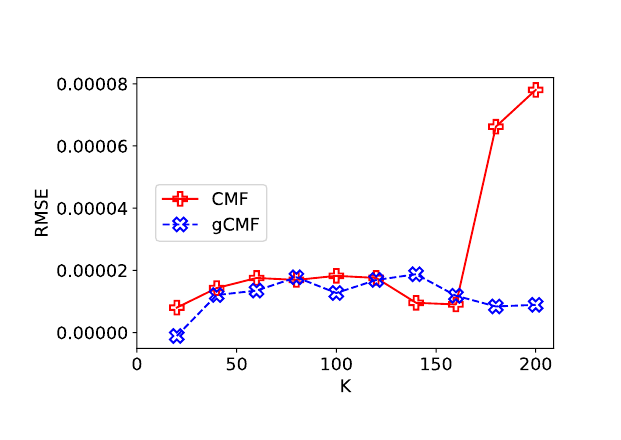}
        }
        \caption{(a) Performance of dCMF, gCMF and  CMF at different values of K (b) Zoomed-in version of (a) where y-axis is y+0.8267, to show performance of CMF and gCMF.}
        \label{fig:simulation_k}
\end{figure*}

We now experimentally study the impact of $K$ (ranging between 20 and 200) on CMF, gCMF and dCMF's performance. 
We generate a synthetic dataset with 4 entities and 3 views based on the recommendation setup (fig. \ref{eg}(b)).
We generated $U^{(e_1)}$, $U^{(e_2)}$, $U^{(e_3)}$, $U^{(e_4)}$ with $K$=100 and the desired dimensions $|e_1|$=400, $|e_2|$=800, $|e_3|$=80 and $|e_4|$=160, with values sampled from a uniform distribution ranging between 0 and 1. We constructed the views $X^{(1)}_{|e_1| \times |e_2|}$, $X^{(2)}_{|e_1| \times |e_3|}$ and $X^{(3)}_{|e_4| \times |e_2|}$ using the corresponding factors, where subscripts indicate dimensions.
For this synthetic dataset, for $K$ = 20,  $\text{p}_\textsubscript{dcmf} \approx$ 134.4K and both $\text{p}_\textsubscript{cmf}, \text{p}_\textsubscript{gcmf} \approx$ 28.8K. For $K$ = 94 , both $\text{p}_\textsubscript{cmf}, \text{p}_\textsubscript{gcmf} \approx$ 135K. 
Thus, the model complexity of CMF and gCMF with $K \approx 94$ can be considered as roughly equivalent to dCMF with $K$ = 20. Similarly model complexity of CMF and gCMF with $K \approx 187$ can be considered as roughly equivalent to that of dCMF with $K$ = 40.

%
%
%

For each $K$ varying between 20 and 200 in steps of 20, we obtained the matrix $X^{{(1)}^\prime}$ using the factors $U^{(1)}$ and $U^{(2)}$ obtained using dCMF, CMF and gCMF. The RMSE between the predicted $X^{{(1)}^\prime}$ and original $X^{(1)}$ is shown in fig. \ref{fig:simulation_k}(a). The RMSE values of CMF and gCMF are nearly the same and hence indistinguishable in the figure; so, a zoomed-in version is shown in fig. \ref{fig:simulation_k}(b).

It can be seen that dCMF consistently outperforms both CMF and gCMF at all values of $K$. 
In particular, we can compare the performance of dCMF at $K=20$ ($40$) and CMF/gCMF at $K=100$ ($200$) that are of roughly equal model complexity and observe that dCMF performs better.
In fact, the performance of CMF or gCMF does not improve with increase in $K$. 

\end{document}